\pdfoutput=1
\documentclass[11pt]{article}

\usepackage[]{acl}

\usepackage{times}
\usepackage{soul}

\usepackage{breakurl}
\usepackage{url}
\makeatletter
\g@addto@macro\UrlSpecials{\do\!{\newline}}
\usepackage{CJKutf8}

\usepackage{color}
\usepackage{graphicx}
\usepackage{amsmath}
\usepackage{booktabs}
\usepackage{algorithm}
\usepackage{algorithmic}
\usepackage{booktabs} % For formal tables
\usepackage{amsthm} 
\usepackage{multirow}
\usepackage{latexsym}
\usepackage{subcaption}

\usepackage[T1]{fontenc}
\usepackage[utf8]{inputenc}

\usepackage{microtype}

\title{Early Discovery of Disappearing Entities in Microblogs}

\author{
  Satoshi Akasaki \\
  The University of Tokyo \\
  \And
  Naoki Yoshinaga\\
  Institute of Industrial Science,\\ 
  the University of Tokyo\\
  \texttt{\{akasaki, ynaga, toyoda\}@tkl.iis.u-tokyo.ac.jp}
  \And
  Masashi Toyoda\\
  Institute of Industrial Science,\\
  The University of Tokyo\\
  }

\newtheorem*{dfn:ec}{\textbf{Emerging contexts}}
\newtheorem*{dfn:pc}{\textbf{Prevalent contexts}}
\newtheorem*{dfn:dc}{\textbf{Disappearing contexts}}
\newtheorem*{dfn:ee}{\textbf{Emerging entities}}
\newtheorem*{dfn:de}{\textbf{Disappearing entities}}
\newtheorem*{dfn:pe}{\textbf{Prevalent entities}}
\newtheorem*{dfn:le}{\textbf{Long-tail entities}}
\newtheorem*{dfn:he}{\textbf{Homographic entities}}

\begin{document}
\maketitle
\begin{abstract}
We make decisions by reacting to changes in the real world, in particular, the emergence and disappearance of impermanent entities such as events, restaurants, and services. 
Because we want to avoid missing out on opportunities or making fruitless actions after they have disappeared, it is important to know when entities disappear as early as possible.
We thus tackle the task of detecting disappearing entities from microblogs, whose posts mention various entities, in a timely manner.
The major challenge is detecting uncertain contexts of disappearing entities from noisy microblog posts.
To collect these disappearing contexts, we design time-sensitive distant supervision, which utilizes entities from the knowledge base and time-series posts, for this task to build large-scale Twitter datasets\footnote{We will release the datasets (tweet IDs) used in the experiments to promote reproducibility.} for English and Japanese.
To ensure robust detection in noisy environments, we refine pretrained word embeddings of the detection model on microblog streams of the target day.
Experimental results on the Twitter datasets confirmed 
the effectiveness of the collected labeled data and refined word embeddings; more than 70\% of the detected disappearing entities in Wikipedia are discovered earlier than the update on Wikipedia, and the average lead-time is over one month.
\end{abstract}

\section{Introduction}\label{disappear:sec:intro}
Our daily actions depend on the state of the real world and its changes, especially changes in the entities available to us. Among the various changes, the beginning of entities, or emerging entities such as new songs, movies, and events, are useful for understanding the trends in interests. 
At the same time, users need to be made aware of the end or disappearance of entities, such as closing stores or discontinuing services, as soon as possible so that they can avoid missing out on opportunities or taking fruitless actions after the entities are no longer available. 
It is also important to collect these entities to maintain knowledge bases (KBs) in which information about the entities is accumulated. 
Studies on discovering out-of-KB or emerging entities~\cite{lin2012, farber2016, wu2016, akasaki2019a} have been successful to an extent, as most of emerging entities have distinct names and can be characterized by mentions to their unseen names. 
In contrast, disappearing entities are not clearly characterized by their mentions, since they continue to be mentioned even after they disappear. 

\begin{table}[t]
\centering
\footnotesize

\begin{tabular}{p{0.95\linewidth}}
\toprule
I'm so \textit{sad to hear} that \textbf{Dave Laing} has \textit{died}. Dave was a very accomplished music industry journalist. But...
\\ \midrule
%Legendary Emmy Award - winning and former NBC4 anchor, \textbf{Doug Adair}, \textit{passed away} peacefully alongside family on Monday in Pleasanton
%\\ \midrule
%ESPN is \textit{shuttering} \textbf{ESPN Deportes Radio} this fall in what appears to be yet another cost cutting move.
%\\ \midrule
%\textit{RIP} \textbf{Ed Corne} What a Physique for under 200lbs! This guy knew how to pose If schwarzenegger was impressed, you…
%\\ \midrule
%Nike \textit{shuts down} Oregan Project ! Less than two weeks after the USADA handed a four - year \textit{ban} to \textbf{Nike Oregon Project} coach
%\\ \midrule
%\textbf{Family Circle}, a pillar of women's magazines, will \textit{shut down} after 87 years.
%\\ \midrule
Can't believe \textbf{Google+} is being \textit{shut down}. It's like when they just pulled Google Friends Connect all over again...
\\ \midrule
%\textbf{Green Mountain College} \textit{Announces Plan To Close} This Spring, Court Rules State Has To Refund Fee For...
%\\ \midrule
%Bernie Sanders on the \textit{planned closure} of \textbf{Hahnemann University Hospital} in Philadelphia: "In the midst of a healthcare crisis …
%\\ \midrule
RT: Here's your \textit{Demolition} Day Planner for \textbf{Martin Tower}. A brief, stray shower can't be ruled out before...
\\ \midrule
%The \textbf{Newseum} will \textit{close} at the end of 2019 following the sale of its building to. Organization says …
%\\ \midrule
\textbf{Red Bull Air Race World Championship} \textit{will not continue} after 2019. URL
\\ \midrule
%We had a blast \textbf{BronyCon} and we're \textit{sad} that it's the \textit{last one} for them. We hope all their staff have a great future …
%\\ \midrule
\textbf{Pristin} to \textit{disband} after 2 years promoting as a group + K - Netz express how raged they are towards Pledis Ent.
\\ \bottomrule
\end{tabular}
\caption{Tweets about disappearing entities (bold) with expressions suggesting their disappearance (italic).}
\label{disappear:table:disappearingentities}
\end{table}

Given such a situation, we take on the new task of discovering disappearing entities from microblogs where news and personal experiences are widely shared. 
To detect the entities' disappearance, we exploit the specific expressions that people use when mentioning disappearing entities in microblogs (Table~\ref{disappear:table:disappearingentities}).
By capturing these contexts, we can discover a variety of entities in the early stage even before they disappear.
To develop a dataset of disappearing entities and contexts, we use time-sensitive distant supervision~\cite{akasaki2019a}, which collects specific contexts of entities by utilizing KB entities and timestamps of microblogs.
Because this method requires the timing of the desired contexts, we extract the year of disappearance for each entity described in Wikipedia and incorporate it into the distant supervision.

We train a named entity recognition (NER) model on the collected entities and contexts to discover disappearing entities. 
However, the NER model performs poorly for microblogs~\cite{derczynski2017}  where posts are short and noisy, and the training data collected by the distant supervision contains noise, making it difficult to train a reliable model. 
We address this issue by considering that multiple posts are likely to mention the target entity when it disappears in the real world.
Concretely, we utilize these posts to refine pretrained word embeddings and incorporate them into the NER model. 
This enables the model to consider the tokens that frequently appear among multiple posts and to recognize disappearing entities robustly.

We built large-scale English and Japanese datasets from Twitter using the proposed time-sensitive distant supervision method.
The experimental results demonstrated that the proposed method outperformed the baseline which simply collected the latest burst of posts about the disappearing entities as the disappearing contexts using time-sensitive distant supervision and used them to train the NER model. 
In addition, the evaluation of relative recall indicated that our method successfully found more than 70\% of the target disappearing entities in Wikipedia. 
Except for entities like persons, whose Wikipedia articles were updated with little or no delay, our method detected entities such as services, facilities, and events on average more than 100 days earlier than the update of the disappearance in Wikipedia.

\section{Definition of Disappearing Entity}
\label{disappear:sec:definition}
In this section, we define the meaning of the term \textit{disappearing entity} in this study.
We consider the entities' disappearance to be the disappearance of its existence from the real world or official announcement of its discontinuation, with reference to the list of ending entities in Wikipedia.\footnote{\url{https://en.wikipedia.org/w/index.php?title=Category:Endings_by_year&from=2000}}

We also refer to \citet{akasaki2019a}, who reported that emerging entities have a specific process from their first appearance to when they become known to the public. 
During this process, they are referred to with specific expressions, \textit{i.e., }emerging contexts. 
Similar to this, we define disappearing entities and contexts by focusing on the fact that specific expressions indicating plans and signs of disappearance appear in contexts not only at the time of disappearance but also in the process leading up to that time as follows:

\begin{dfn:dc}
Contexts in which the writers assumed the readers do not know the disappearance of the entities.
\end{dfn:dc}
\begin{dfn:de}
Entities still being observed in disappearing contexts.
\end{dfn:de}

Table~\ref{disappear:table:disappearingentities} lists examples of disappearing entities and contexts. 
By properly identifying these disappearing contexts, we can detect corresponding disappearing entities in the early stages even before they actually disappear. This is especially important for entities such as stores that are closing or services and events that are ending so that users can take action before they are gone.\footnote{Note that how early the disappearance is detected depends on the entity type.} We later confirm the solidness of these definitions by evaluating the inter-rater agreement of disappearing entities acquired from microblogs and by demonstrating that the disappearing entities can be detected before the update of the disappearance in Wikipedia (\S~\ref{disappear:subsec:results}).

\section{Related Work}
\label{disappear:sec:related}
To our knowledge, there have been no studies that detect disappearing entities in a timely manner, which is the target of our work. We briefly review the current studies related to our task. 

\subsection{Entity Extraction}
Because existing entity related tasks such as NER~\cite{nadeau2007, lample2016, akbik2018, akbik2019} and entity linking~\cite{shen2014, kolitsas2018, martins2019} do not take into consideration whether the entities have already disappeared from the real world, it is difficult to detect disappearing entities by using these techniques.

As a counterpart to disappearing entities, \citet{akasaki2019a} aimed to find newly emerging entities.\footnote{Although there are other studies that detect emerging entities like \citet{hoffart2014} and \citet{wu2016}, they actually target out-of-KB entities.}
They focused on the fact that people use expressions that suggest novelty when mentioning emerging entities and defined the entities based on these expressions (contexts). 
They proposed a distant supervision method called time-sensitive distant supervision to collect emerging contexts efficiently utilizing KB entities and microblog timestamps and then developed a NER model using constructed data to detect emerging entities.
Although this is the contrasting task to ours, the same method cannot be directly because of the difference in contexts handled in this study.

\subsection{Temporal Event Extraction}
As part of information extraction, several researchers~\cite{ritter2012, nguyen2015, lu2017, liu2020} have tackled the task of extracting an event (\textit{e.g., }birth of a person) and its predefined attributes and arguments (such as time and places) from the text.
Although it is possible to use this technique to detect disappearing entities, they are defined as a supervised task based on manual annotation~\cite{doddington2004,aguilar2014} and only a few entity types such as `person' or `military conflict'  are chosen for the current annotation. It is thus unrealistic to extend manual annotations to detect various entity types.

KBP2011~\cite{ji2010,mcclosky2012} introduced the temporal slot filling task, in which the duration of an event is identified from the given text, entities (\textit{e.g., }Steve Jobs), and their events (\textit{e.g., }become CEO).
The events are attributes defined in Freebase, and they include the disappearance of certain entities such as when a person dies. However, the types of entities handled in this task are also only a few (`person' and `organization'). 
Moreover, although target entities are given in advance for this task, we have to detect mentions of the entities in the settings of microblogs.

\begin{figure}[t]
    \flushleft
    \includegraphics[width=0.49\textwidth]{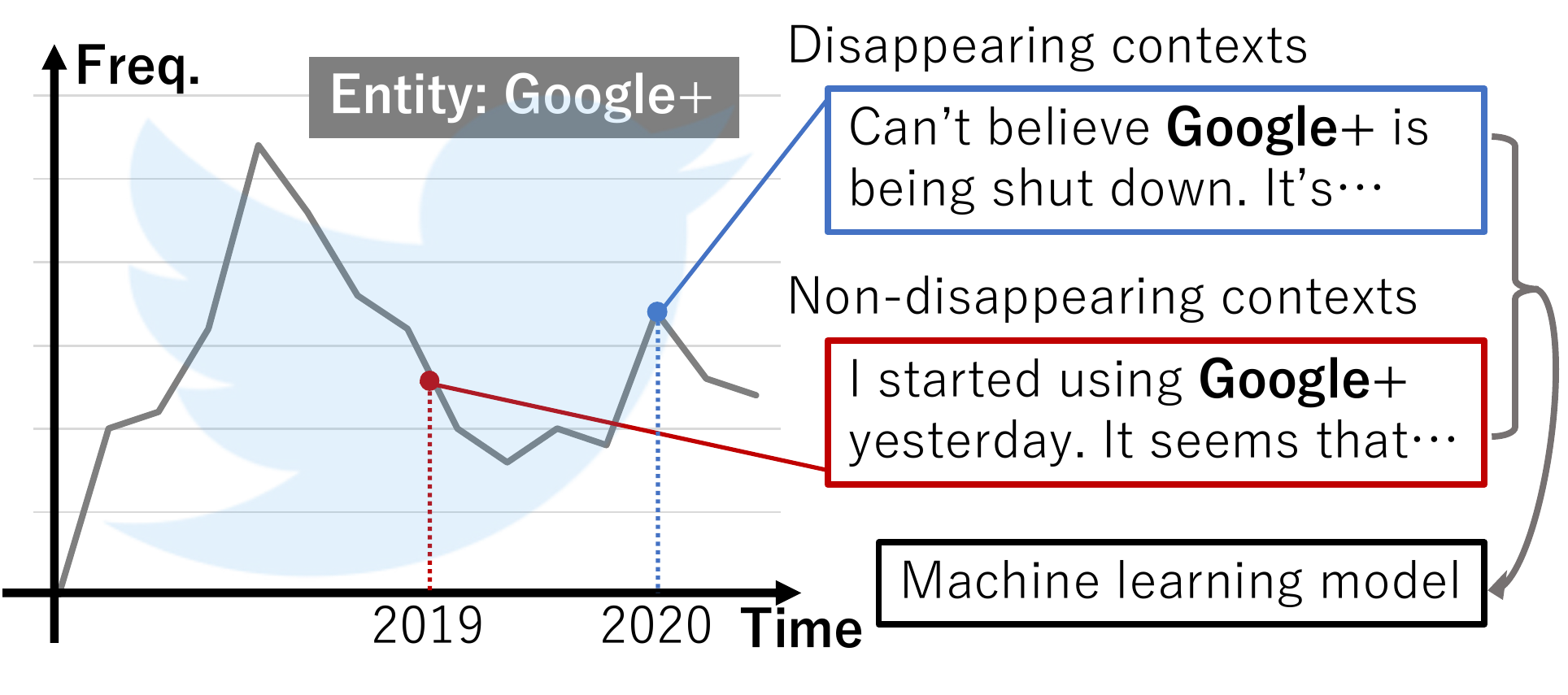}
    %\vspace*{-0.3cm} 
    \caption{Time-sensitive distant supervision: For the entities retrieved from a KB, disappearing contexts and other contexts are collected from microblogs by utilizing the year of entities' disappearance. We then train a sequence labeling model using the obtained contexts.}
    \label{disappear:fig:proposedds}
\end{figure}

\begin{figure*}[t]
    \centering
    \includegraphics[width=0.85\textwidth]{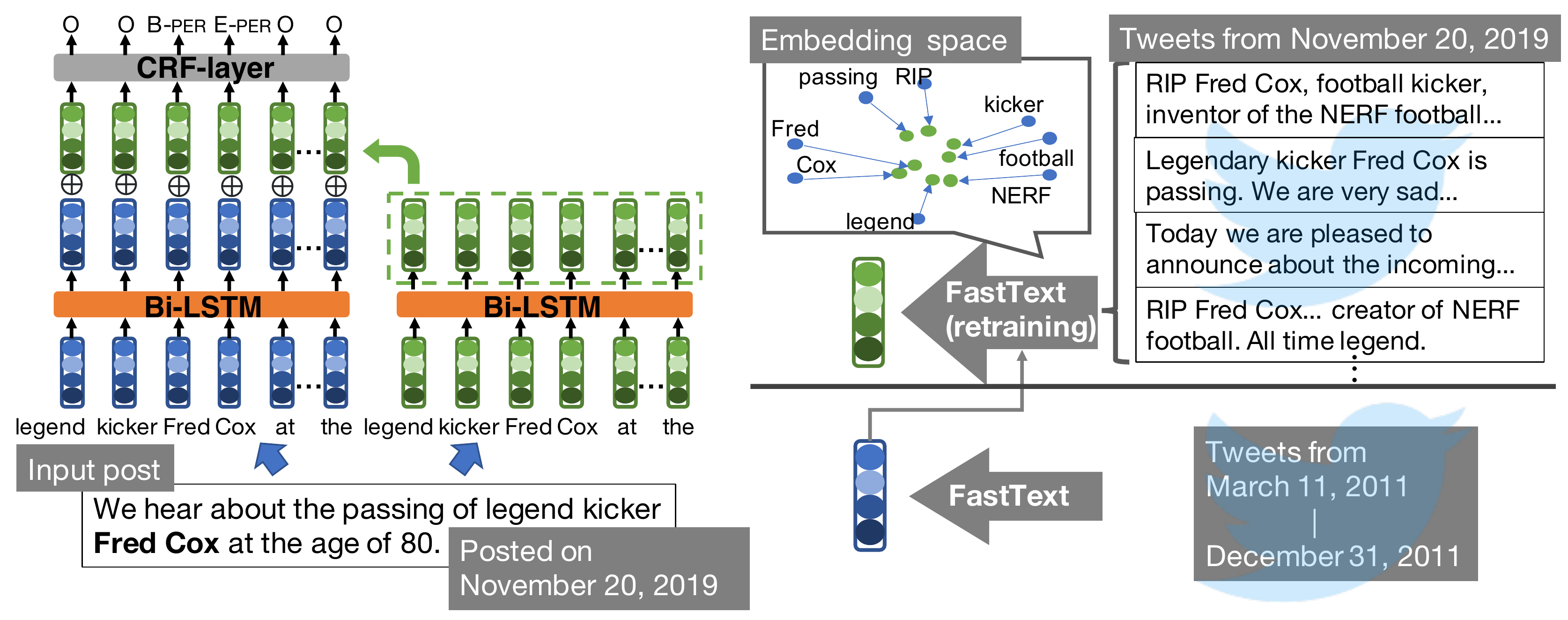}
    %\vspace*{-0.3cm} 
    \caption{Sequence labeling with refined word embeddings: We refine pretrained word embeddings using the Twitter stream on the day of the input post, and feed them into the NER model for robust training and prediction.}
    \label{disappear:fig:proposedword}
\end{figure*}

\section{Proposed Method}
\label{disappear:sec:proposal}
The objective of our proposed method is to discover disappearing entities in microblogs.
We target Twitter, where various sources including news articles and personal posts, are extensively and instantly shared. 
To accurately collect contexts of disappearing entities~(\S~\ref{disappear:sec:definition}), we extract the timings of the entities' disappearance described in Wikipedia to elaborate time-sensitive distant supervision, which is originally designed for detecting emerging entities~\cite{akasaki2019a}, for our task.
To ensure that an detection model can make robust predictions using the noisy dataset constructed using distant supervision and for noisy microblog posts, we refine pretrained word embeddings to acquire features from multiple occurrences of disappearing entities and feed them into the model.

\subsection{Time-sensitive Distant Supervision}
\label{disappear:subsec:distant_supervision}
While \citeauthor{akasaki2019a} were able to gather contexts of emerging entities by collecting the posts when they first appeared using the timestamps from microblogs, it is difficult to directly apply this implementation of time-sensitive distant supervision to disappearing entities because the timing of their disappearance is not clear. 
Therefore, we explicitly feed the timing of the entities' disappearance extracted from Wikipedia to time-sensitive distant supervision to collect disappearing entities and their contexts more accurately (Figure~\ref{disappear:fig:proposedds}). The specific procedure is as follows:

\paragraph{Step 1. Collecting disappearing entities}
We first collect candidates of disappearing entities while excluding noise.
To collect only entities that have actually disappeared, we refer to the list of ending entities in Wikipedia\footnotemark[1] and gather the titles of articles, categories, and their year of disappearance. 
We excluded entities whose year of their first appearance on Twitter was the same as the year of their disappearance as they could be emerging entities and contaminate the contexts. 
We also remove entities that have the ambiguity page so that the contexts are not contaminated by homographic entities which share the same namings with other entities (\textit{e.g.,} ``Go'' can refer to a programming language, a board game, or a verb).

To acquire entity types, we manually map the category of the article to the coarse-grained type;\footnote{We used this method to obtain coarse-grained types as there are few mappings of disappearing entities in DBpedia that are defined as \citet{akasaki2019a}.} for example, the entity `Daft Punk' is mapped to the type `\textsc{Group}.' 

\paragraph{Step 2. Collecting disappearing contexts}
In contrast to emerging entities, where the first appearance of the entity is often the emerging context, the latest posts of disappearing entities contain noisy unrelated contexts because they continue to be mentioned in microblogs even after they have disappeared. 
Therefore, for each collected entity, we utilize the year of disappearance and frequency of appearance on Twitter to gather their disappearing contexts. 
Specifically, we randomly collect $k$ posts of the day with the highest number of occurrences in the given year, assuming that the timing that received the most attention in the year of the disappearance includes the disappearing contexts.

For each collected entity, \citet{akasaki2019a} collected contexts that differed from the positive examples as negative examples to avoid overfitting the NER model when detecting mentions of positive examples. 
Thus, we similarly collected random $k$ non-disappearing contexts as negative examples from posts prior to the year in which we collected positive examples for each entity. 
This enables the model to discriminate between disappearing contexts and other contexts.

We adopt the BILOU scheme~\cite{ratinov2009} for the NER tags; we label the disappearing entities in the positive examples with BILU and their entity types, and label the rest with O.

\subsection{Finding Disappearing Entities}
\label{disappear:subsec:sequence_labeling}
We train an NER model for finding disappearing entities using the collected data. 
Because we target short and noisy microblog posts in this study and build the dataset using distant supervision, it is difficult to train the model stably. 
Thus, we focus on the fact that multiple posts mentioning the disappearance of the entities often appear when entities disappear from microblogs. 
By obtaining features from these posts, we can enhance the training and prediction of the model even with noisy microblog posts. 
To do this, we propose an unsupervised method of refining pretrained word embeddings as features from multiple posts of a Twitter stream. We also devise a sequence labeling model for finding disappearing entities using the refined word embeddings as additional input (Figure~\ref{disappear:fig:proposedword}).

\subsubsection{Refining Pretrained Word Embeddings}
Our objective is to extract features from multiple posts in the Twitter stream. 
However, because the surface of the target entity is unknown at the test time of NER, it is difficult to collect only relevant posts of that entity from the massive Twitter stream. 
To address this issue, we use, as the additional embedding layers of our NER model, pretrained word embeddings further fine-tuned on the posts on the day of detecting disappearing entities, together with the original pretrained word embeddings.
This enables the refined word embeddings to reflect the tokens and their co-occurrences in the Twitter stream of the target day without the need for post selection. 
The specific procedure is as follows:

First, we train the base word embeddings $v_{base}$ using the posts prior to the period in which we collected the data in \S~\ref{disappear:subsec:distant_supervision}. 
We use fastText~\cite{bojanowski2017}, a method of constructing word embeddings, to deal with unknown words.

Next, we use the Twitter stream from each date $d$ of the posts collected in \S~\ref{disappear:subsec:distant_supervision} to retrain the word vector $v_{base}$ for obtaining $v_{d}$. The resulting $v_{d}$ can be interpreted as capturing the temporary semantic change of $v_{base}$ on date $d$, and it can be treated as an auxiliary input of models for various tasks.

\subsubsection{NER with Refined Word Embeddings}
We adopt \citet{akbik2019} model, which uses long short-term memory with conditional random field (LSTM-CRF)~\cite{huang2015} and flair embeddings~\cite{akbik2018}.\footnote{We do not use models pretrained on large corpora such as BERT because they do not consider timestamps and the data may contain future data, which leads to unrealistic settings.} 
This model inputs character embeddings, which are encoded by the pretrained character-based bidirectional-LSTM language model, and pretrained word embeddings into the word-based bidirectional-LSTM and makes predictions through the CRF layer.

Based on this model, we prepare another word-based bidirectional-LSTM and input the refined word embeddings $v_{d}$ corresponding to the date of the input post. 
We then concatenate the hidden layers of each LSTM at each time and feed them into the CRF layer. 
This enables the model to consider the global information of the given Twitter stream other than the input post.

\begin{table*}[t]
\tiny
\fontsize{7pt}{9pt}\selectfont
    \centering
        \begin{tabular}{@{\,}l@{\quad}l@{\,\,\,}r@{\,\,\,}r@{\quad}l@{\,}}
            \toprule
            \multicolumn{2}{@{\,}l@{\,\,\,}}{\textbf{TYPE}} & \textbf{\# entities} & \textbf{\# posts} & \textbf{examples of disappearing context (truncated)}\\
            &\textbf{Wikipedia categories}\\
            \midrule
\multicolumn{2}{@{\,}l@{\,\,\,}}{\textsc{Person}} & \textbf{780} & \textbf{24689}\\
& Deaths & 780 & 24689 & \textbf{Roger Ailes} died of complications of a subdural hematoma after he fell at home, hit his head. \\
&&&& …to ski one day with Olympic Legend \textbf{Stein Eriksen} at Deer Valley. He passed yesterday at…\\
%&&&& \textbf{Billy Brewer} has passed away. A wonderful guy who loved Coaching \& was a personable guy…\\
&&&& RIP \textbf{Dave Rosenfield}: @USER executive, International League schedule maker, one-time…URL \\
%&&&& The passing of Dr. \textbf{Irving Moskowitz} is a tremendous loss for the world, the Jewish people…\\
%&&&&  \\
\midrule
\multicolumn{2}{@{\,}l@{\,\,\,}}{\textsc{Creative work}} & \textbf{975} & \textbf{24222}\\
& American\_television\_series & 381 & 12413 & RT @USER: See what's coming up in the \textbf{Diggnation} Finale airing next week…\\
& British\_television\_series & 139 & 3650  & Wait…is this episode supposed to be the \textbf{Metalocalypse} finale or the start…\\
%& Korean\_television\_series & 75 & 2039  & Late Friday news dump: Comedy Central has canceled ``\textbf{Not Safe with Nikki Glase}''…\\
& Web\_series & 57 & 1679 & Final Series of \textbf{McLevy} this week on @USER The Scotland office are missing cast \& crew …\\
%& Philippine\_television\_series & 56 & 875 & …didnt see it coming but still frustrated as all out that \textbf{Tekzilla} is now cancelled…\\
%& Canadian\_television\_series & 44 & 570  & Last Episode of \textbf{Star Trek Continues} was released today "To Boldly Go: Part II"…\\
& Others (36 types) & 398 & 6480 & Crying while watching the last episode of \textbf{Packed to the Rafters} :( such a great tv show \\
\midrule
\multicolumn{2}{@{\,}l@{\,\,\,}}{\textsc{Location}} & \textbf{240} & \textbf{3545}\\
& Buildings\_and\_structures & 54 & 949 & @USER: After 67 years \textbf{Clemson House} is gone in seconds. (This version is sped up) URL\\
& Educational\_institutions & 36 & 647 & \textbf{Coleman University} closing its doors after loss of accreditation URL\\
%& Sports\_venues & 34 & 598 & \textbf{Baylor's Floyd Casey Stadium} is no more. May you and your tarp rest in peace…URL \\
& Restaurants & 24 & 286 & …food news of HASH Everyone loves Lynn's. \textbf{Lynn's Paradise Cafe} abruptly serves its final meal…\\
%& Populated\_places & 23 & 232 & BREAKING: Macy's will close its \textbf{Landmark Mall} store as part of its swath of 2017 closures:\\
& Others (20 types) & 126 & 1663 & Sad news, crime sleuths. The \textbf{National Museum of Crime and Punishment} in D.C. will close at the…\\
\midrule
\multicolumn{2}{@{\,}l@{\,\,\,}}{\textsc{Group}} & \textbf{1187} & \textbf{30384}\\
& Musical\_groups & 453 & 11559 & …in Flyleaf anymore. Adam's not in Three Days Grace anymore. \textbf{My Chemical Romance} broke up…\\
& Retail\_companies & 79 & 3187 & The convenience store chain, \textbf{My Local}, is to be placed in administration,9 months after it was sold…\\
& Airlines & 60 & 1535 & The second hand car giant \textbf{Carcraft} has gone bust, with the loss of around 500 jobs across the UK…\\
%& Political\_parties & 56 & 654 & The \textbf{Australian Democrats} officially deregistered by AEC bc less than 500 members. - ABC news…\\
%& Mass\_media\_companies & 44 & 1351 & \textbf{Mad Catz} files for bankruptcy and is ceasing operations URL URL\\
& Others (68 types) & 595 & 14103 & \textbf{The Foreign Policy Initiative}, a right-leaning foreign-policy think tank, will cease operations…\\
\midrule
\multicolumn{2}{@{\,}l@{\,\,\,}}{\textsc{Event}} & \textbf{186} & \textbf{4940}\\
& Sporting\_events & 111 & 2492  & The \textbf{Adidas Grand Prix} in NY, replaced by Rabat meet in Diamond League series, is set to become…\\
& Events & 33 & 1462 & Live music blow. MT @TomTilley: C3 confirms \textbf{Big Day Out} will NOT go ahead in 2015. \\
%& Awards & 20 & 646 & \textbf{World Series Formula V8 3.5} to end after 2017 season URL URL \\
& Sports\_leagues & 16 & 137 & The cancellation of \textbf{Women's Professional Soccer League} in US is bad news for England \& Team…\\
%& Music\_festivals & 6 & 203 & \textbf{Camden Crawl} organisers issue full statement following liquidation, blaming poor ticket sales:… \\
& Others (2 types) & 26 & 849 & \textbf{The Foreign Policy Initiative}, a right-leaning foreign-policy think tank, will cease operations…\\

\midrule
\multicolumn{2}{@{\,}l@{\,\,\,}}{\textsc{Service\&Product}} & \textbf{709} & \textbf{17574}\\
& Magazines & 187 & 2998 &  was what inspired me to write for mags. RT @USER: Future closes \textbf{Nintendo Gamer magazine} \\
& Internet\_properties & 131 & 5916  & Google is going to shut down \textbf{Orkut} on September 30….Haven't seen that site since 2007 I… \\
& Products\_and\_services & 119 & 3450  & \textbf{The PlayStation 3} has ended production according to the official PlayStation Japan website.\\
%& Television\_stations & 105 & 1727  & If you get past the exciting times and digital strategy, the closure of \textbf{STV2} is four paragraphs…\\
%& Publications & 53 & 1043  & Joe weighs in on the demise of \textbf{The Tampa Tribune} and the changing Bucs beat. (full story) URL\\
& Others (10 types) & 272 & 5210  & Just heard that \textbf{Yorkshire Radio} is no more. Great shame as @USER was pretty much the voice of …\\
\midrule

\multicolumn{2}{@{\,}l@{\,\,\,}}{\textsc{Total}} & \textbf{3213} & \textbf{81850}\\
\bottomrule
        \end{tabular}
\caption{Statistics of English disappearing entities and disappearing contexts in training data 
obtained from our Twitter archive by time-sensitive distant supervision.}
\label{disappear:table:dseng}
\end{table*}

%% test data
\begin{table}[htbp]
\begin{tabular}{ll}
\scriptsize

\begin{minipage}[h]{0.5\hsize}
%\captionsetup{width=.70\textwidth}
%\begin{flushleft}
        \begin{tabular}{@{}l@{\quad}l@{\,\,}r@{\,\,}r@{\,\,}l@{}}
            \toprule
            \multicolumn{2}{@{}l@{}}{\textbf{TYPE}} & \textbf{\#ent.} & \textbf{\#posts} 
            &\\
            \midrule

\multicolumn{2}{@{}l@{}}{\textsc{Person}} & \textbf{147} & \textbf{422}\\
% & Deaths & 147 & 422 \\
% \midrule
\multicolumn{2}{@{}l@{}}{\textsc{Creative work}} & \textbf{10} & \textbf{23}\\
% & American\_television. & 6 & 16 \\
% & Radio\_programme & 2 & 3 \\
% %& Web\_series & 1 & 2 \\
% %& Philippine\_television. & 1 & 2 \\
% & Others (2 types) & 2 & 4 \\
% \midrule
\multicolumn{2}{@{}l@{}}{\textsc{Location}} & \textbf{46} & \textbf{113}\\
% & Sports\_venues & 7 & 19 \\
% & Shopping\_malls & 5 & 15 \\
% %& Restaurants & 5 & 10 \\
% %& Railway\_lines & 2 & 5 \\
% %& Populated\_places & 2 & 2 \\
% & Others (11 types) & 34 & 84 \\
% \midrule
\multicolumn{2}{@{}l@{}}{\textsc{Group}} & \textbf{103} & \textbf{270}\\
% & Retail\_companies & 12 & 33 \\
% & Game\_companies & 4 & 12 \\
% %& Transport\_companies & 4 & 11 \\
% %& Telecom.\_companies & 1 & 2 \\
% %& Religious\_org. & 1 & 3 \\
% & Others (28 types) & 87 & 225 \\
% \midrule
\multicolumn{2}{@{}l@{}}{\textsc{Event}} & \textbf{8} & \textbf{19}\\
% & Sporting\_events & 4 & 7 \\
% % & Sports\_leagues & 3 & 9 \\
% % & Recurring\_events & 1 & 3 \\
% & Others (2 types) & 4 & 12 \\
% \midrule
\multicolumn{2}{@{}l@{}}{\textsc{Service\&Product}} & \textbf{43} & \textbf{114}\\
% & Television\_stations & 5 & 12 \\
% & Publications & 4 & 10 \\
% %& Radio\_stations & 2 & 4 \\
% %& YouTube\_channels & 1 & 3 \\
% %& Space\_probes & 1 & 3 \\
% & Others (9 types) & 34 & 92 \\
\midrule
\multicolumn{2}{@{}l@{}}{\textsc{Total}} & \textbf{357} & \textbf{961}\\
\bottomrule
\label{typing:table:dseng_test}
        \end{tabular}
%        \end{flushleft}
\vspace{-10pt}
\subcaption{English}
\end{minipage}

\begin{minipage}[h]{0.5\hsize}
\scriptsize
%\captionsetup{width=.0\textwidth}
%\begin{flushleft}
        \begin{tabular}{@{}l@{\quad}l@{\,\,}r@{\,\,\,}r@{\,\,}l@{}}
            \toprule
            \multicolumn{2}{@{}l@{}}{\textbf{TYPE}} & \textbf{\#ent.} & \textbf{\#posts}
            &\\
            \midrule

\multicolumn{2}{@{}l@{}}{\textsc{Person}} & \textbf{73} & \textbf{220}\\
% & Deaths & 73 & 220 \\
%& \begin{CJK}{UTF8}{ipxm}没\end{CJK}& 73 & 220 \\
% \midrule
\multicolumn{2}{@{}l@{}}{\textsc{Location}} & \textbf{42} & \textbf{114}\\
% & \begin{CJK}{UTF8}{ipxm}施設\end{CJK}& 23 & 64 \\
% & \begin{CJK}{UTF8}{ipxm}鉄道駅\end{CJK}& 10 & 26 \\
% & \begin{CJK}{UTF8}{ipxm}建築物\end{CJK}& 7 & 18 \\
% & \begin{CJK}{UTF8}{ipxm}博物館\end{CJK}& 1 & 3 \\
% & \begin{CJK}{UTF8}{ipxm}スポーツ施設 \end{CJK}& 1 & 3 \\
% & Buildings & 23 & 64 \\
% & Stations & 10 & 26 \\
% & Structures & 7 & 18 \\
% & Museums & 1 & 3 \\
% & Sports\_venues & 1 & 3 \\
% \midrule
\multicolumn{2}{@{}l@{}}{\textsc{Group}} & \textbf{64} & \textbf{173}\\
% & \begin{CJK}{UTF8}{ipxm}音楽グループ\end{CJK}& 29 & 84 \\
% & \begin{CJK}{UTF8}{ipxm}企業\end{CJK}& 26 & 68 \\
% & \begin{CJK}{UTF8}{ipxm}組織\end{CJK}& 7 & 17 \\
% & \begin{CJK}{UTF8}{ipxm}スポーツチーム\end{CJK}& 2 & 4 \\
% & Musical\_groups & 29 & 84 \\
% & Companies & 26 & 68 \\
% & Organizations & 7 & 17 \\
% & Sports\_teams & 2 & 4 \\
% \midrule
\multicolumn{2}{@{}l@{}}{\textsc{Event}} & \textbf{9} & \textbf{25}\\
% & \begin{CJK}{UTF8}{ipxm}イベント\end{CJK}& 5 & 14 \\
% & \begin{CJK}{UTF8}{ipxm}スポーツイベント\end{CJK}& 4 & 11 \\
% & Events & 5 & 14 \\
% & Sporting\_events & 4 & 11 \\
% \midrule
\multicolumn{2}{@{}l@{}}{\textsc{Service\&Product}} & \textbf{47} & \textbf{131}\\
% & \begin{CJK}{UTF8}{ipxm}オンラインゲーム\end{CJK}& 31 & 89 \\
% & \begin{CJK}{UTF8}{ipxm}雑誌\end{CJK}& 11 & 29 \\
% & \begin{CJK}{UTF8}{ipxm}ウェブサイト\end{CJK} & 5 & 13 \\
% & Online\_games & 31 & 89 \\
% & Magazines & 11 & 29 \\
% & Websites & 5 & 13 \\
 \midrule
\multicolumn{2}{@{}l@{}}{\textsc{Total}} & \textbf{235} & \textbf{663}\\
\bottomrule
        \end{tabular}
%        \end{flushleft}
\vspace{6pt}
\subcaption{Japanese}
\end{minipage}
\end{tabular}
\caption{Statistics of disappearing entities and disappearing contexts in test data.}
\label{disappear:table:dsengja_test}
\end{table}

\section{Experiments}
\label{disappear:sec:experiment}
We performed our task of discovering disappearing entities using datasets built from Twitter.

\subsection{Data}
\label{disappear:subsec:data}
We constructed the Twitter datasets by time-sensitive distant supervision.
We targeted English and Japanese, which are the top two languages used on Twitter~\cite{alshaabi2021}, and use our archive of Twitter posts that were retrieved\footnote{Timelines of 26 popular Japanese users starting from 2011 have been continuously collected using user\_timeline API, while the user set has been iteratively expanded to those who were mentioned or retweeted by existing users.} using Twitter APIs. 
The archive consists of more than 50B posts (32\% are English and 20\% are Japanese; this does not deviate much from the actual data~\cite{alshaabi2021}). 
We tokenized each post using spaCy (ver.~2.0.12)\footnote{\url{https://spacy.io}} with the en\_core\_web\_sm model for English and MeCab (ver.~0.996)\footnote{\url{https://taku910.github.io/mecab}} with ipadic (ver.~2.7.0) for Japanese. We removed URLs, usernames, and hashtags from the text.

In Step 1 of \S~\ref{disappear:subsec:distant_supervision}, we collected article titles of ending entities in Wikipedia from 2012 to 2019 using the Wikipedia dump from June 20th, 2021. 
We undersampled the types \textsc{Person} and \textsc{Creative work} to 1,000 entities as they are much larger than the other types. 
We then excluded entities as described and carried out Step 2 by setting $k$ to 100, as in~\cite{akasaki2019a}. 

We split the collected data into training data (2012-2018) and test data (2019). 
For the training data, we obtained a total of 163,700 English and 150,204 Japanese tweets, including the same number of disappearing and other contexts for 3,213 English entities and 1,906 Japanese entities, respectively. 
For model selection, we used 10\% of the training data as the development data. 

For the test data, from the collected positive examples of disappearing entities, we randomly selected three posts for each entity and asked three annotators (the first author and two graduate students) to judge whether each context is accompanied by a disappearing context. 
We adopt the positive contexts with the answers agreed upon by two or more annotators. 
Then, for each entity of the resulting data, we asked the annotators to determine the non-disappearing contexts using the collected negative examples of the entity and selected the same number of posts as the positive examples. 
We obtained an inter-rater agreement of 0.722 for English and 0.786 for Japanese by Fleiss's Kappa, both of which show substantial agreement. As a result, we obtained a total of 1,922 English and 1,326 Japanese tweets for 357 English entities and 235 Japanese entities, respectively, as the test data.

The resulting disappearing entities and disappearing contexts of training data (English) and test data (English and Japanese) are shown in Table~\ref{disappear:table:dseng} and~\ref{disappear:table:dsengja_test}, respectively. 
The entity types that are manually categorized into \textsc{Person} and \textsc{Group} account for a large proportion. 
These types of entities, occupied by persons, musical groups, and companies, are more likely to disappear. 
Because we use Wikipedia as the list of disappearing entities, there is a bias in the composed categories, and some categories are absent in certain languages. 
However, we can still detect entities of the missing categories because if the coarse type of entities is the same, their context tends to be somewhat similar regardless of their category (\textit{e.g.,} \textsc{LOCATION} types tend to be mentioned with the term \textit{close}).

\subsection{Models}
\label{disappear:subsec:model}
%The following models were implemented for comparison:
We implemented following models for comparison:

\smallskip\noindent\textbf{Proposed (TDS + RefEmb):}
We implemented LSTM-CRF with flair embeddings~\cite{akbik2018} and proposed refined embeddings (RefEmb) using the training data constructed by the proposed time-sensitive distant supervision (TDS).
We refined pretrained fastText embeddings for each input post using tweets on the day of the input post (about 1 to 2M tweets for each day from 2012 to 2019) and additionally fed them into the LSTM-CRF.

\smallskip\noindent\textbf{Proposed (TDS):}
To verify the effectiveness of refined word embeddings, we removed the part related to RefEmb (green part in Figure~\ref{disappear:fig:proposedword}) from Proposed (TDS + RefEmb) and used with the same training data, optimization, and parameters. 
This method does not consider the multiple posts of the target day when recognizing entities.

\smallskip\noindent\textbf{Baseline:}
To verify the effectiveness of the constructed training data, we collected the latest posts of disappearing entities using the original version of time-sensitive distant supervision, which does not consider the timing of the entities' disappearance.
Specifically, we follow~\citet{akasaki2019a} and for each ending entity in Wikipedia from 2012 to 2018, we collected 100 retweets of the last day (through 2018) in which the entity appeared more than ten times as positive examples (disappearing contexts).
For negative examples, we obtained the same number of posts from more than one year before the date when we collected the positive examples.
By using the collected 25,920 posts for 2,867 entities in English and 6,777 posts for 1,733 entities in Japanese, we trained LSTM-CRF with flair embeddings using the same optimization and parameters as Proposed (TDS).
Because this method does not consider the timing of the entities' disappearance, many noisy contexts may be collected.

\begin{table}[tbp]
\scriptsize
\centering
  \begin{tabular}{l||c} 
    \textbf{Parameter} & \textbf{Value} \\ \hline
    Character embedding size (LM) & 30 \\
    Dimension of Character Bi-LSTM (LM) & 1024 \\
    SGD learning rate (LM) & 20.0 \\
    Batch size (LM) & 100 \\
    Word embedding size (LSTM-CRF) & 300 \\
    Dimension of Word Bi-LSTM (LSTM-CRF) & 256 \\
    Batch size (LSTM-CRF) & 32 \\ 
    Dropout (LSTM-CRF) & 0.5 \\ 
    SGD learning rate (LSTM-CRF) & 0.01 \\ \hline
  \end{tabular}
    \caption{Hyperparameters of character-based language model (LM) and LSTM-CRF.}
  \label{disappear:table:hyperparameters}
\end{table}

\begin{table}[t]
\centering
\footnotesize
%\begin{minipage}{0.75\textwidth}
\begin{minipage}{0.5\textwidth}
%\centering

\begin{tabular}{l||ccc}
    & \textbf{Prec.} & \textbf{Rec.} & \textbf{F$_1$}  \\
                   \hline
Proposed (TDS + RefEmb)   & 0.730      & \textbf{0.671}   & \textbf{0.699}  \\
Proposed (TDS)             & \textbf{0.766}      & 0.587   & 0.665  \\
Baseline (TDS)             & 0.514      & 0.184   & 0.271  \\
\end{tabular}

\subcaption{English}
\end{minipage}

\begin{minipage}{0.5\textwidth}
%\centering

\begin{tabular}{l||ccc}
    & \textbf{Prec.} & \textbf{Rec.} & \textbf{F$_1$} \\
                   \hline
Proposed (TDS + RefEmb)  & \textbf{0.850}      & \textbf{0.599}   & \textbf{0.708}    \\
Proposed (TDS)            & 0.828      & 0.532   & 0.648   \\
Baseline (TDS)            & 0.743      & 0.143   & 0.240    \\
\end{tabular}

\subcaption{Japanese}
\end{minipage}

    \caption{Overall performance of each method for English and Japanese.}
\label{disappear:table:res_all}
\end{table}

\begin{table}[t]
\centering
\footnotesize
%\begin{minipage}{0.75\textwidth}
\begin{minipage}{0.5\textwidth}
\centering

\begin{tabular}{l||ccc}
    & \textbf{Prec.} & \textbf{Rec.} & \textbf{F$_1$}  \\
                   \hline
\textsc{Person}             & 0.865      & 0.901   & 0.883  \\
\textsc{Creative work}      & 0.480      & 0.500   & 0.490  \\
\textsc{Location}           & 0.727      & 0.491   & 0.586  \\
\textsc{Group}              & 0.570      & 0.526   & 0.547  \\
\textsc{Service\&Product}   & 0.566      & 0.409   & 0.475  \\
\textsc{Event}              & 0.800      & 0.444   & 0.571  \\
\end{tabular}

\subcaption{English}
\end{minipage}

\begin{minipage}{0.5\textwidth}
\centering

\begin{tabular}{l||ccc}
    & \textbf{Prec.} & \textbf{Rec.} & \textbf{F$_1$} \\
                   \hline
\textsc{Person}             & 0.948      & 0.858   & 0.901    \\
\textsc{Location}           & 0.814      & 0.569   & 0.670   \\
\textsc{Group}              & 0.818      & 0.418   & 0.553    \\
\textsc{Service\&Product}   & 0.777      & 0.552   & 0.645  \\
\textsc{Event}              & 0.250      & 0.042   & 0.071     \\
\end{tabular}

\subcaption{Japanese}
\end{minipage}

    \caption{Performance of Proposed (TDS + RefEmb) for each coarse type}
\label{disappear:table:res_eachtype}
\end{table}

\subsection{Settings}
\label{disappear:subsec:setting}
\paragraph{Implementation and model parameters} We use Keras~(ver.~2.3.1)\footnote{\url{https://keras.io}} for implementing all the models. 
For flair embeddings, we set hyperparameters as suggested in~\cite{akbik2018} and trained the character-based bidirectional LSTM language model from 2B English tweets for English and 800M Japanese tweets, respectively, both posted from March 11th, 2011 to December 31st, 2011. 
The hyperparameters are listed in Table~\ref{disappear:table:hyperparameters}. 
Using the same tweets, we trained 300-dimensional word embeddings using fastText\footnote{\url{https://fasttext.cc/}} and used them to initialize the embedding layers of LSTM-CRF by concatenating with flair embeddings. 
We optimized all models using stochastic gradient descent and chose the model in the epoch with the highest F$_1$-score on the development data. 

\paragraph{Evaluation methods}
To evaluate the accuracy of our method, we apply the models to each post in the test data constructed in \S~\ref{disappear:subsec:data} and evaluate the results using the CoNLL-2003~\cite{sang2003} schema, which measures precision, recall, and F$_1$-score.

To evaluate the relative recall and the detection immediacy of our method, we follow the experiments designed for emerging entities~\citet{akasaki2019a}. Specifically, for each ending entity in Wikipedia that disappeared in 2019 (2,608 for English and 763 for Japanese), we applied our method to all of the posts in 2019 in which each entity appeared (437,816 for English and 202,666 for Japanese). 
Then we determined how many target entities could be discovered from the posts and how much earlier those entities could be detected before the corresponding Wikipedia articles were categorized as ending.

\subsection{Results and Analysis}
\label{disappear:subsec:results}

\paragraph{Overall accuracy of models}
Table~\ref{disappear:table:res_all} shows the micro precision, recall, and F$_1$-score for all models. Both the proposed methods outperformed the baseline, which collected training data without considering the timing of the entities' disappearance. The performance of the baseline is low because it was trained with noisy data. This shows that our time-sensitive distant supervision successfully collected the disappearing contexts. Our Proposed (TDS + RefEmb) detected the entities with the highest accuracy, which means that the refined word embeddings worked effectively. In particular, the recall was improved in both Japanese and English, indicating that entities that could not be recognized by only using the features of a single post can be successfully detected by utilizing multiple posts.

\paragraph{Detailed accuracy of optimal model}
To analyze the behavior of Proposed (TDS + RefEmb), we show the precision, recall, and F$_1$-score for each type in Table~\ref{disappear:table:res_eachtype}.
The accuracy of \textsc{Person} type entities is high in both Japanese and English. This is likely because the numerous entities of this type exist in the training data and the person's names themselves are easy to recognize from the surface. The \textsc{Creative work} type is not present in Japanese, and the accuracy for that type is low in English because the disappearance of these entities is uncertain for the nature of the type. For example, even when the final episode of a television drama airs on TV, it remains in various other media. This causes the training data to be contaminated with diverse contexts, resulting in the model's low accuracy. The accuracy for \textsc{Event} type entities is the lowest in Japanese. 
This is likely because the number of training data is small and thus the model could not be sufficiently trained.

\paragraph{Relative recall and detection immediacy}

\begin{table}[tbp]
\scriptsize
%\begin{tabular}{cc}

\begin{minipage}[t]{0.5\textwidth}
%\fontsize{6.5pt}{9pt}\selectfont
%\captionsetup{width=1.0\textwidth}
\begin{center}
%    \centering
    % ynaga dirty latex
        \begin{tabular}{@{\,}l@{\,\,\,}l@{\,}r@{\,\,\,\,}r@{\ }r@{\,}r@{\!\!\!\!\!\!\!\!\!}r@{\,}}
            \toprule
             \multicolumn{2}{@{\,}l}{\textbf{TYPE}} & \textbf{\# entities} & \multicolumn{2}{@{\,}r}{\textbf{\# found (\%)}} & \multicolumn{2}{@{\,}r}{\textbf{lead-days}} \\
            &  & & & & %multicolumn{2}{@{\,}l@{\,}}{\textbf{mean (median)}} \\_
            \multicolumn{1}{@{\,}c@{\!\!\!}}{\textbf{mean}} & \multicolumn{1}{c@{\,}}~{\textbf{(median)}} \\
            \midrule
\multicolumn{2}{@{\,}l}{\textsc{Person}} & 1838 & 1668 & (90.75\%) & 21 & (0) \\
\multicolumn{2}{@{\,}l}{\textsc{Creative work}} & 351 & 123 & (35.04\%) & 173 & (64) \\
\multicolumn{2}{@{\,}l}{\textsc{Location}} & 73 & 38 & (52.05\%) & 150 & (46) \\
\multicolumn{2}{@{\,}l}{\textsc{Group}} & 163 & 99 & (60.74\%) & 195 & (131) \\
\multicolumn{2}{@{\,}l}{\textsc{Service\&Product}} & 93 & 47 & (50.54\%) & 172 & (91) \\
\multicolumn{2}{@{\,}l}{\textsc{Event}} & 25 & 8 & (32.00\%) & 85 & (35) \\
\multicolumn{2}{@{\,}l}{\textsc{Unmapped}} & 65 & 34 & (52.31\%) & 237 & (171) \\
\midrule
\multicolumn{2}{@{\,}l}{\textsc{Total}} & 2608 & 2017 & (77.34\%) & 49 & (0) \\
\multicolumn{2}{@{\,}l}{\textsc{Total}} (w/o \textsc{Person}) & 770 & 349 & (45.32\%) & 139 & (77) \\
\bottomrule
        \end{tabular}
        \end{center}
        %\caption{Relative recall and time advantage over entity types of English disappearing entities detected with Proposed (TDS + RefEmb).}
        \subcaption{English}
    \label{disappear:table:recdetail_eng}
\end{minipage}

%\hfill

\begin{minipage}[t]{0.5\textwidth}
%\fontsize{6.5pt}{9pt}\selectfont
%\captionsetup{width=.90\textwidth}
\begin{center}
%    \centering
    % ynaga dirty latex
        \begin{tabular}{@{\,}l@{\,\,\,}l@{\,}r@{\,\,\,\,}r@{\ }r@{\,}r@{\!\!\!\!\!\!\!\!\!}r@{\,}}
            \toprule
             \multicolumn{2}{@{\,}l}{\textbf{TYPE}} & \textbf{\# entities} & \multicolumn{2}{@{\,}r}{\textbf{\# found (\%)}} & \multicolumn{2}{@{\,}r}{\textbf{lead-days}} \\
            &  & & & & %multicolumn{2}{@{\,}l@{\,}}{\textbf{mean (median)}} \\_
            \multicolumn{1}{@{\,}c@{\!\!\!}}{\textbf{mean}} & \multicolumn{1}{c@{\,}}~{\textbf{(median)}} \\
            \midrule
\multicolumn{2}{@{\,}l}{\textsc{Person}} & 515 & 446 & (86.60\%) & 31 & (0) \\
\multicolumn{2}{@{\,}l}{\textsc{Location}} & 48 & 31 & (64.58\%) & 149 & (117) \\
\multicolumn{2}{@{\,}l}{\textsc{Group}} & 121 & 56 & (46.28\%) & 181 & (147) \\
\multicolumn{2}{@{\,}l}{\textsc{Service\&Product}} & 63 & 43 & (68.25\%) & 154 & (98) \\
\multicolumn{2}{@{\,}l}{\textsc{Event}} & 16 & 6 & (37.50\%) & 183 & (171) \\
\midrule
\multicolumn{2}{@{\,}l}{\textsc{Total}} & 763 & 582 & (76.28\%) & 63 & (0) \\
\multicolumn{2}{@{\,}l}{\textsc{Total}} (w/o \textsc{Person}) & 248 & 136 & (54.84\%) & 132 & (101) \\
\bottomrule
        \end{tabular}
        \end{center}
        %\caption{Relative recall and time advantage over entity types of Japanese disappearing entities detected with Proposed (TDS + RefEmb).}
        \subcaption{Japanese}
    \label{disappear:table:recdetail_jp}
\end{minipage}

%\end{tabular}
        \caption{Relative recall and time advantage over entity types of English and Japanese disappearing entities detected with Proposed (TDS + RefEmb).}
        \label{disappear:table:recdetail}
\end{table}

Table~\ref{disappear:table:recdetail} shows the distribution of the types of target entities, detection ratio, and lead time against the Wikipedia update time for both languages. Overall, Proposed (TDS + RefEmb) detected 2,017 (77.34\%) English and 582 (76.28\%) Japanese disappearing entities. 
More than 80\% of \textsc{Person} entities were detected in both languages, while the other types were found 45.32\% for English and 54.84\% for Japanese. 
Note that some target entities are low frequency in our Twitter archive and do not appear in disappearing contexts. Because our method utilizes such disappearance signals as clues, it is difficult to discover entities that are mentioned without disappearing contexts. 
This is the current limitation of our method.

For detection immediacy, we confirmed that 77.34\% of the discovered English entities (1,560 out of 2,017) and 85.39\% of the discovered Japanese entities (497 out of 582) were detected earlier than their update in Wikipedia.
Entities that were slow to be detected were often either mentioned infrequently with disappearing contexts or their  updates were unusually fast.
The mean (and median) lead days of the first day when our method detected each entity against their update date were 49 (and 0.13) for English and 63 (and 0.44) days for Japanese.
In particular, for other types of entities other than person, the lead days were 139 days (and 77) for English and 132 days (and 101) for Japanese. 
This demonstrates the detection immediacy of our method. 
Our method could discover entities such as Durgin-Park (restaurant), Rolling Acres Mall (shopping mall), and CiteULike (Internet property), which are the types of entities whose disappearance is important to know in advance. 
This should prevent users from making fruitless actions or missing out on opportunities because the average lead time for these types is more than 100 days. 
Interestingly, the updates for \textsc{Person} type entities in Wikipedia are faster for both languages. 
Since this kind of uncontrolled disappearance (death) occurs suddenly, it is difficult to detect these entities before they disappear.
It also shows that only certain types of entities are updated faster in Wikipedia.

\section{Conclusion}
We introduced the task of discovering disappearing entities in microblogs (\S~\ref{disappear:sec:intro}, \S~\ref{disappear:sec:definition}, and \S~\ref{disappear:sec:related}).
To deal with the uncertainty of entity disappearance, we considered the year of disappearance in Wikipedia and fed it into time-sensitive distant supervision (\S~\ref{disappear:subsec:distant_supervision}).
To perform the detection from noisy microblog posts, we proposed the method of refining pretrained word embeddings using the Twitter stream (\S~\ref{disappear:subsec:sequence_labeling}).
Experimental results demonstrated that our method outperformed the baseline method and successfully found more than 70\% of the target disappearing entities in Wikipedia and they were detected more than a month earlier than the update of the disappearance in Wikipedia (\S~\ref{disappear:subsec:results}).

We plan to extract information such as the specific time of disappearance to make the discovered entities more useful for applications.

\section*{Ethical Considerations}
The dataset was collected using Twitter's official API\footnote{\url{https://developer.twitter.com/en/docs/twitter-api}} and in compliance with Twitter's terms of use.
Only the tweet IDs of the tweets used in the experiments will be made public, and we ensure that their redistribution is in compliance with Twitter's developer policy.\footnote{\url{https://developer.twitter.com/en/developer-terms/agreement-and-policy}}
Researchers cannot collect deleted tweets or tweets of private users, thus protecting user privacy.

\bibliography{acl}
\bibliographystyle{acl_natbib}

\end{document}